\documentclass{article}
\usepackage{graphicx}
\usepackage{cite}
\usepackage{amsmath,amsfonts,amssymb}
\usepackage{url}
\usepackage{array,booktabs}
\usepackage{wrapfig}
\usepackage{pbox}
\usepackage{multirow}
\usepackage{hhline}
\usepackage{lscape}
\usepackage{rotating}
\usepackage[bottom]{footmisc}

\newcommand{\f}{\mathbf{f}}
\newcommand{\x}{\mathbf{x}}
\newcommand{\argmin}{\text{argmin}}
\newcommand{\lc}{\mathcal{L}_c}
\newcommand{\ls}{\mathcal{L}_s}
\newcommand{\pool}[1]{\text{\path{pool #1}}}
\newcommand{\conv}[1]{\text{\path{conv #1_1}}}

\graphicspath{{fig/inputs/}{fig/results/}}

\begin{document}
\title{Content-Aware Neural Style Transfer}
\author{Rujie Yin}

\maketitle

\section*{Motivation}
In the emerging inter-disciplinary field of art and image processing, algorithms have been developed to assist the analysis of art work. In most applications, especially brush stroke analysis, high resolution digital images of paintings are required to capture subtle patterns and details in the high frequency range of the spectrum. Algorithms have been developed to learn styles of painters from their digitized paintings to help identify authenticity of controversial paintings. However, high quality testing datasets containing both original and forgery are limited to confidential image files provided by museums, which is not publicly available, and a small sets of original/copy paintings painted by the same artist, where copies were deferred to two weeks after the originals were finished. Up to date, no synthesized painting by computers from a real painting has been used as a negative test case, mainly due to the limitation of prevailing style transfer algorithms. 

There are two main types of style transfer algorithms, either transferring the tone (color, contrast, saturation, etc.) of an image, preserving its patterns and details, or distorting the texture uniformly of an image to create ``style". In this paper, we are interested in a higher level of style transfer, particularly, transferring a source natural image (e.g. a photo) to a high resolution painting given a reference painting of similar object. The transferred natural image would have a similar presentation of the original object to that of the reference painting. In general, an object is painted in a different style of brush strokes than that of the background, hence the desired style transferring algorithm should be able to recognize the object in the source natural image and transfer brush stroke styles in the reference painting in a {\it content-aware} way such that the styles of the foreground and the background, and moreover different parts of the foreground in the transferred image, are consistent to that in the reference painting. 

Recently, an algorithm based on deep convolutional neural network has been developed to transfer artistic style from an art painting to a photo\cite{Gatys2015c}. Successful as it is in transferring styles from impressionist paintings of artists such as Vincent van Gogh to photos of various scenes, the algorithm is prone to distorting the structure of the content in the source image and introducing artifacts/new content in the background of the transferred image. We investigate conditions and methods to improve this neural style transfer algorithm to be content-aware along with the goal of synthesizing a high resolution painting in more realism style. 

\section*{Background}
Deep convolutional neural network (CNN) has been extensively applied to various image processing following its initial success in image classification of ImageNet\cite{Deng09imagenet:a}. A CNN typically consists of a sequence of sets of convolutional layers followed by non-linear rectifier and local pooling. Like classical image processing, each convolutional layer performs linear spatial filtering on the input image or the output feature maps from the previous layer; the local pooling down-samples spatial information similar to dyadic down-sampling in classic wavelet decomposition of images. On the other hand, the non-linearity introduced by the rectifier enables CNN to approximate complicated non-linear transforms, mapping images to a feature space where different classes of images become linearly separable. The depth of CNN is proportional to the capacity of the function space that can be approximated, hence a deeper CNN is more ``expressive" than a shallower one, encoding more feature information of image classes.

\subsection*{Neural Style Transfer}
Due to the richness of features that a deep CNN can possess, pre-trained CNN models on massive datasets like ImageNet with high accuracy in image classification have been used as feature mappings from image domain to abstract feature spaces for other image processing. In particular, the neural style transfer algorithm in \cite{Gatys2015c} is based on VGG-Network(VGG-Net)\cite{simonyan2014very}, a near-human performance deep CNN model in image recognition. 

Given an input image $\x$, let $\f_{l,k}(\x)$ be the output $N_l$ dimensional feature map\footnote{this is the same as the term {\it filter response} used in\cite{Gatys2015c}} of VGG-Net at the $l^{th}$ layer and  the $k^{th}$ filter, where $N_l$ is the size of the corresponding spatial grid, then the neural style transfer is formulated as the following optimization problem,
\begin{align}\label{eq: style-opt}
\x^* = \argmin_{\x'} &\sum_{l \in \lc}\sum_{ k}\big\Vert\, \f_{l,k}(\x_c)-\f_{l,k}(\x')\,\big\Vert^2_2 \notag\\
&+ \lambda \sum_{l\in\ls}w_l\sum_{i, j}\big\vert\, \f_{l,i}^\top\, \f_{l,j}(\x_s) -\f_{l,i}^\top\,\f_{l,j}(\x')\,\big\vert^2
\end{align}
where $\x_c$ is the source (content) image, $\x_s$ is the reference painting (style image) and $\x^*$ is the style transferred source image. The first term in \eqref{eq: style-opt} penalizes the difference between the feature vectors of source image and transferred image at a set of layers $\lc$ in VGG-Net, such that the content of these two images are close. To measure the closeness of styles, the second term involves the covariance of feature maps $\f_{l,\cdot}$ corresponding to different filters at a fixed layer $l$ in the set $\ls$ weighted by $w_l$. This formulation is also used in another closely related neural algorithm for texture synthesis\cite{Gatys2015b}, where only a texture (style) image is used to generate a synthetic image presenting similar texture. In the style transfer algorithm, the sets of constraint layers are chosen as $\lc:$ \texttt{conv4\_2} and $\ls^{\text{style}}:$ \path{conv1_1,conv2_1,conv3_1,conv4_1,conv5_1}. In the texture synthesis algorithm, $\ls^{\text{txtr}}:$ \path{conv1_1,pool1,conv2_1,pool2,conv3_1,pool3,conv4_1,pool4}. We defer the comparison of these two settings of constraint layers to the discussion of methods in the next section.

\section*{Methods}
We propose a content-aware neural style transfer algorithm based on the framework of \eqref{eq: style-opt}. We first present the base algorithm for the simple case where the source content image and the reference style image are of the same resolution. The algorithm is then generalized to the case where the content image is in lower resolution than the style image.
The output of our algorithm is a synthesized painting that could have been painted by the same painter of the reference painting from the content image. We always assume that the content image (photo) and the style image (painting) contain similar objects to minimize the ambiguity of how the generated painting should look like.

\subsection*{Style Layers $\ls$}
For each layer $l$ in the network, let $s_l = N_{l-1}/N_l$ be the scaling factor of that layer. For a convolutional layer, the scaling factor is 1; for a pooling layer, the scaling factor is its kernel size. Therefore, the scale of features characterized at level $l$ is $\sigma_l = s_1\cdots s_l$. Specifically, a convolutional layer has the same feature scale as the last pooling layer ahead of it, and the final pooling layer has the maximum feature scale in a network. 
It is observed in both \cite{Gatys2015b} and \cite{Gatys2015c} that a texture image can be synthesized by fitting the covariance of feature maps $\f_{l,\cdot}$ at layer $l\in\ls$, namely minimizing the second term in \eqref{eq: style-opt} with random initialization. Moreover, the generated texture image is spatially uniform and it represents similar texture of the reference style image at the scales of style layers $\ls$. 
In style transfer \cite{Gatys2015c} and texture synthesis \cite{Gatys2015b}, different style layers  $\ls$ are used, but the scales of these layers are the same, namely $1, \sigma_{\pool{1}}, \sigma_{\pool{2}},\sigma_{\pool{3}}$ and $\sigma_{\pool{4}}$. Using layers at different scales collaboratively enforces a hierarchical structural constraint on the texture in the synthesized image. Given a reference style image whose texture is at scale $\sigma_s$, that is its energy concentrated within a circle of radius $\sigma_s^{-1}$ centered at the origin of the frequency spectrum, $\ls$ should contain layers at scale larger than $\sigma_s$ to retain full texture pattern.

Although restricting $\ls$ to layers at smaller scales  than $\sigma_s$ results in less satisfactory result, the statistics of the reference style pattern at finer scales are still preserved in the generated image. In fact, a style pattern at scale $\sigma_s$ consists of sub-patterns at finer scales, such as color dots, single or group of brush strokes, etc.  Therefore, preserving  statistics of $\ls = \{\text{\path{conv1_1}}\}$ results in a synthesized style image consists of color dots, where the portion of areas in different colors is roughly the same as that in the reference image, see results in \cite{Gatys2015c} and \cite{Gatys2015b}. More generally, the sub-patterns at scale $\sigma_l, l\in\ls$ are preserved in the synthesized image, which is a {\it re-arrangement} of the composition of reference style image at unconstrained scales. Together with the fact that VGG-Net is not rotation invariant,  this insight of re-arrangement leads to an important condition on the input content image and the style image for content-aware style transfer: the object in content image and that in style image should be in the same orientation and in the same relative scale to image size, which is also numerically validated in section Experiments.

\subsection*{Content Layers $\lc$ and Intialization}
In contrast to collaborative multi-scale style layer constraints, a single deep layer is typically used in content constraint. As discussed in \cite{Gatys2015c}, the output feature maps from a shallow layer in VGG-Net is highly redundant and sufficient to reconstruct the input image. The similarity between the reconstruction and the input image decreases as the layer whose output feature maps are constrained becomes deeper. Choosing a shallow content layer (e.g. \path{conv2_1}) thus does not provide much freedom in transferring the content image into a new style, whereas choosing a very deep content layer (e.g. \path{conv5_1}) does not penalize big distortion of object in the synthesized image. In \cite{Gatys2015c}, \path{conv4_2} is used as content layer constraint together with random initialization of the optimization \eqref{eq: style-opt}.

This choice of content layer works well for indefinite style transfer where the style and the content are almost independent. As the weight $\lambda$ before the style penalty term in \eqref{eq: style-opt} changes, the result synthesized image varies from presenting highly similar object in the content image with minimal fusion of the reference style to showing only the reference style without the knowledge of content. However, simply adjusting $\lambda$ does not provide a synthesized image that retains adequate information in both content and style, which is required in content-aware style transfer.

To resolve this problem in content-aware style transfer, we initialize the optimization algorithm with an image close to the content image instead of a random image. It is well known that optimization on deep neural network like \eqref{eq: style-opt} using stochastic gradient descent achieves local minimum hence starting from an image close to the content image puts a strong prior on the final optimization result. On the other hand, this change of initialization degrades the style constraint as the potential barrier between the initial point and the set of points (images) $\mathcal{I}_s$ in the reference style is higher than that between $\mathcal{I}_s$ and a random point. In numerical experiments, we observe that solving \eqref{eq: style-opt} with fixed $\lambda$, and the same $\ls,\lc$ as that used in \cite{Gatys2015c}, the synthesized image is much more similar to the content image if initialized with the content image than with random initialization. Therefore, we enhance the style constraint and weaken the content penalty in \eqref{eq: style-opt} by setting $\ls:\conv{1},\pool{1},\conv{2},\pool{2},\conv{3},\pool{3},\conv{4},\pool{4},\conv{5} $ and $\lc :\text{\path{conv5_2}}$, where $\ls$ contains layers at all scales and $\lc$ contains one layer at coarsest scale. As $\lambda \rightarrow \infty$, \eqref{eq: style-opt} degenerated to the texture synthesis formulation. As long as we don't initialize randomly and keep a mild penalization on content in the objective function, the result image fully combines style and content from the inputs.

\section*{Experiments}
In this section, we demonstrate style transfer algorithms with different configurations of style layers $\ls$, content layers $\lc$, optimization weight parameter $\lambda$, initialization and input alignment. We first show results on same-resolution style transfer, where the style pattern of the input style image is in almost the same resolution as that of the content image. Based on the best configuration for content-aware style transfer in this easier case, we then discuss its extension to the super-resolution case, where the style pattern of the input style image is in much higher resolution than that of the content image. 

\subsection*{Data and Preprocessing}

\begin{figure}
\includegraphics[width=.49\textwidth]{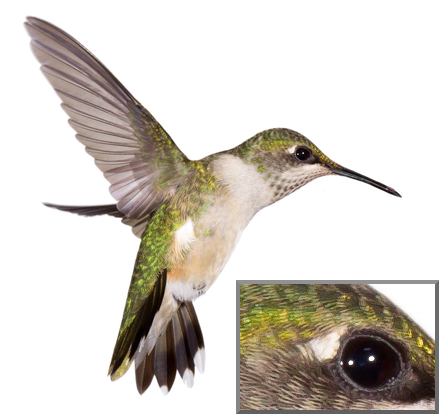}\hfill
\includegraphics[width=.45\textwidth]{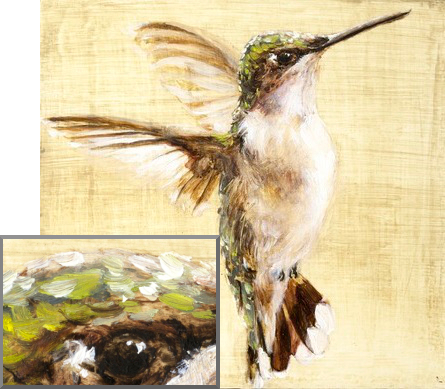}
\caption{Left: content image, a photo of humming bird with white background (1896 $\times$ 1864 pixel), right: reference painting, a humming bird painted on a square wood board (2024 $\times$ 1947 pixel)}
\label{fig: input}
\end{figure}

In the following numerical experiments, a pair of content image and style image shown in Fig.\ref{fig: input} have been used. The style image is a high resolution scan of a real painting of a humming bird by Charlotte Caspers,\footnote{ this painting is selected from a collection of paintings painted by Charlotte Caspers in 2012, served as a public research dataset for digital art authenticity analysis. The dataset is available at \url{http://services.math.duke.edu/~rachel/resources/charlotte2012.html}} and it's available to download at \url{http://services.math.duke.edu/~rachel/charlotte_new2/Nr2_original.tiff}. The content image was manually found through Google search, where the foreground humming bird is in similar type and pose to that in the painting. The content photo is originally1896 $\times$ 1864 pixel and the painting is rescaled  to  2024 $\times$ 1947 from 4049 $\times$ 3894, such that it has the same resolution as the content image. We use the implementation of the style transfer algorithm in \cite{Gatys2015c} on Github \cite{neuralstyle}, which allows flexible configuration of the original algorithm including the one we propose.

\begin{wrapfigure}{l}{0.45\textwidth}
\centering
\includegraphics[width = .45\textwidth]{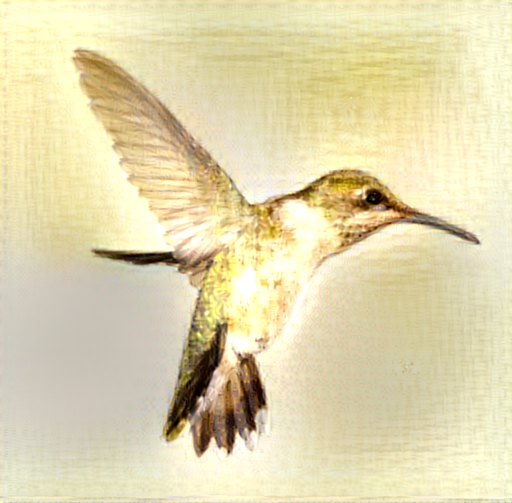}
\caption{synthesized image by basic style transfer using the content image and style image in Fig.\ref{fig: input} (both input images are down-sampled to 512 $\times$ 503 pixel)}
\label{fig: full-transfer}
\vspace{-10pt}
\end{wrapfigure}

Due to the size of the input content image and style image and the limitation of our computation resource, it is impossible to directly take the content image and style image as input and synthesize a style transferred image. To understand the potential problems of applying the original style transfer algorithm to the full content image and style image, we run the algorithm on input images of decreased size (512 $\times$ 503). The result is shown in Fig.\ref{fig: full-transfer}, where the background is not well synthesized and the color of the body of the humming bird is not close to that in the painting. These problems are not related to the resolution of the input images, but they are closely related to the initialization as we will show below, as we will show in experiments of same-resolution style transfer.

\begin{figure}
\includegraphics[width=.49\textwidth]{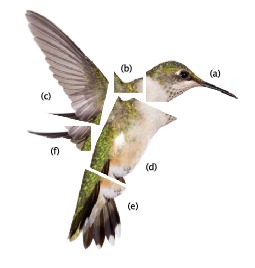}\hfill
\includegraphics[width=.49\textwidth]{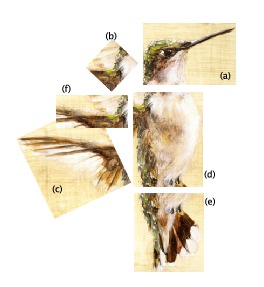}
\caption{cropped corresponding parts from content image and style image in Fig.\ref{fig: input}: (a) head, (b) nape, (c) full wing, (d) body, (e) tail, (f) second wing.  }
\label{fig: parts}
\end{figure}

In order to generate synthesized image of the same resolution of the input image, the content image is segmented into major parts such as head, body, tail and wings, see Fig. \ref{fig: parts}. Each part overlaps with its adjacent parts such that  a full-size style transferred image can be reconstructed by merging the synthesized image on parts. The foreground object in the painting is segmented into parts and accordingly the content image. The segmented parts in both images are in the same relative spatial configuration except that for part (f), the secondary wing in the content image, is matched with a sub-region of of part (c), the major wing in the style image for better consistency. We thus focusing on content-aware style transfer on each part instead of the full image in our experiments.

\subsection*{Same-resolution Style Transfer}
The goal of same-resolution style transfer is to transfer the style pattern from the style image in its original resolution to the content image. Moreover, when the style is content-dependent, e.g. the brush stroke and wood grain pattern in background is different from the brush stroke used to depict foreground object, the algorithm should transfer the background pattern to the background of the content image and the foreground style to the foreground.

\begin{figure}
\includegraphics[width = \textwidth]{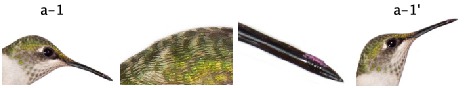}
\includegraphics[width = \textwidth]{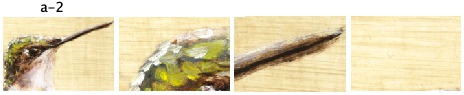}\\[.2em]
\caption{a-1: part (a) of content image, a-2: part (a) of style image, a-1': rotated a-1 aligned in the same direction of a-2; second column: a zoomed in patch at the upper left of the head; third column: a zoomed in patch of the beak; fourth column in the bottom row: a zoomed in background patch in a-2. }
\label{fig: input-p1}
\end{figure}

We first show the necessity of input alignment and non-random initialization. Consider transferring style on part (a), the head of the bird. a-1 and a-2 in Fig.\ref{fig: input-p1} are cropped part (a) from the full content image and style image respectively. The difference of detailed features between a-1 and a-2 are shown in the middle columns of Fig.\ref{fig: input-p1}. The painting doesn't capture very fine details in the photo, such as the pink lump on the tip of the beak and the vein of feathers on the head. Besides the local dissimilarity in texture, the two input images are not globally aligned in the same direction. For input with alignment, we use a-1' in Fig.\ref{fig: input-p1} as the content image together with a-2 in our experiment.

\begin{figure}
\includegraphics[width = \textwidth]{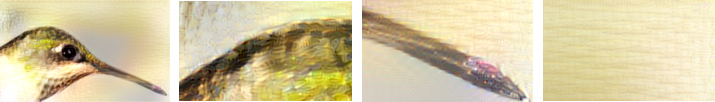}\\[.2em]
\includegraphics[width=  \textwidth]{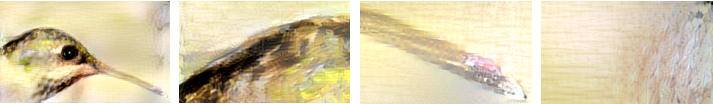}\\[.2em]
\includegraphics[width=  \textwidth]{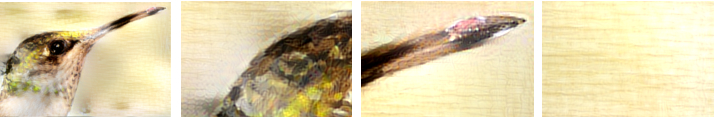}\\[.2em]
\includegraphics[width= \textwidth]{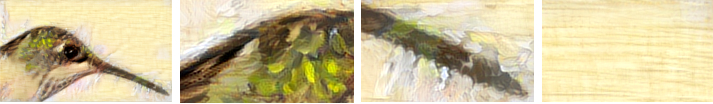}\\[.2em]
\includegraphics[width=  \textwidth]{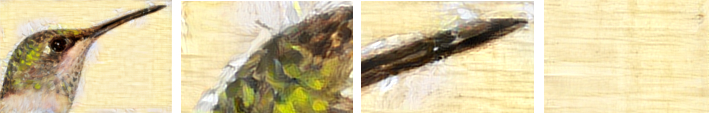}\\
\begin{center}
\renewcommand{\arraystretch}{1.5}
\begin{tabular}{ | >{\centering}p{6em}| >{\centering}p{5em}| >{\centering}p{4em} | >{\centering}p{1.5em} | >{\centering}p{5.5em}|>{\centering\arraybackslash}p{5em}|} 
\hline
configuration & style layers  & content layers  & $\lambda$& initialization & input     alignment\\
\hline\hline
I & $\ls^{\text{style}}$ & \path{conv4_2} & 20 & random & no\\
\hline
II & $\ls^{\text{txtr}}$ & \path{conv4_2} & 20 & random & no\\
\hline
III &  $\ls^{\text{txtr}}$ & \path{conv4_2} & 20 & random & yes\\
\hline
IV &  $\ls^{\text{txtr}}$  $+\, \conv{5}$ & \path{conv5_2} & 200 & content image & no\\
\hline
V & $\ls^{\text{txtr}}$  $+\, \conv{5}$ & \path{conv5_2} & 200 & content  image & yes\\
\hline
\end{tabular}
\end{center}
\caption{Figures shown in columns from left to right: synthesized image from part (a) in the content photo and the reference painting, a zoomed in patch at the upper left of the head, a zoomed in patch of the beak, a zoomed in background patch in the upper right (the first, second and fourth rows)/bottom right (the third and fifth rows) corner.  The results in rows from top to bottom correspond to configurations I to V in the table above.}
\label{fig: Nr2_p1}
\end{figure}

We consider the following four configurations solving \eqref{eq: style-opt}: (i) basic style transfer, i.e. the original algorithm in \cite{Gatys2015c}, (ii) texture enhanced style transfer, where the style layers $\ls^{\text{txtr}}$ in \cite{Gatys2015b} is used instead of $\ls^{\text{style}}$, (iii) texture enhanced with input alignment, which is the same as (ii) but use a-1' instead of a-1 as content image, (iv) content-aware style transfer without input alignment, (v) content-aware style transfer, our proposed configuration. Fig.\ref{fig: Nr2_p1} shows the results and a table of these four configurations.

As shown in Fig.\ref{fig: Nr2_p1}, the results with random initialization, the first to third rows, suffer from un-even lightening in the background despite of the difference in style constraint and input alignment. This is because that the covariance of feature maps, which encodes style information, is estimated globally across the full image in \eqref{eq: style-opt} not distinguishing the foreground and the background. As the style penalty term is not content-aware, the style of the foreground and the background in the synthesized image can be mixed. 
In particular, the background of the painting is uniform of small variance, and that of the synthesized image has larger variance due to the constraint on variance of color channels in style layers $\ls$ mainly contributed by the foreground object in the painting. 
Therefore, the spatial information contained in the initial point is a strong regularization, and the results initialized with the content images have uniform lightening in the backgrpund as in the style image.

Furthermore, the results with input alignment (i.e. using a-1' and a-2 as input), the third and fifth rows, have better brush stroke simulation than those without alignment (i.e. using a-1 and a-2 as input). In particular, the direction of brush strokes of the beak aligns in the same direction of the beak when the inputs are aligned, but not when mis-aligned, see the third column of Fig.\ref{fig: Nr2_p1}. This is due to the fact that VGG-Net is not rotation-invariant and the same object with different orientations are encoded in different filter channels in the layers of the network, especially in the shallow layers, where feature information is not highly aggregated. When the content is consistent with the content image, it cannot be consist with the style image, so are their feature maps in shallow layers, unless both input images are in the same orientation. On the other hand, the shallow layers cannot be dropped from $\ls$ for reconstruction of texture in fine scale.

In addition, including pooling layers in $\ls$ helps to enhance textures from the style image and diminish fine details from the content image in the synthesized image. See the first and second rows in Fig.\ref{fig: Nr2_p1} for comparison of basic configuration and the texture enhanced configuration.


\begin{figure}
\includegraphics[width = \textwidth]{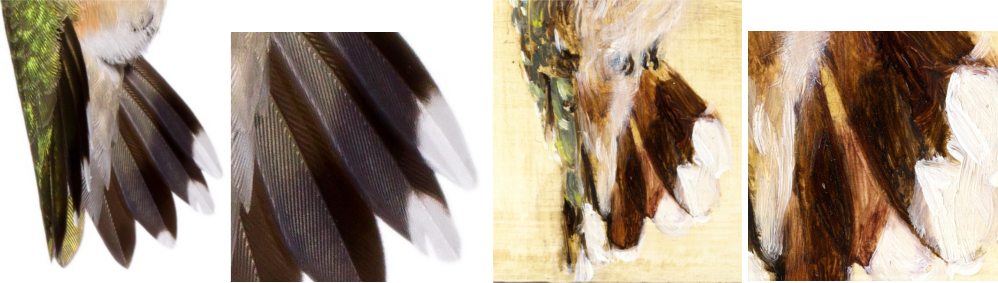}\\[.4em]
\includegraphics[width = \textwidth]{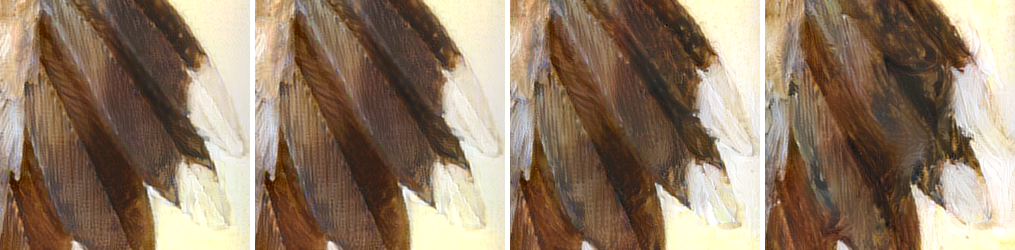}\\
\begin{center}
\renewcommand{\arraystretch}{1.5}
\begin{tabular}{ | >{\centering}p{6em}| >{\centering}p{5em}| >{\centering}p{4em} | >{\centering}p{1.5em} | >{\centering}p{5.5em}|>{\centering\arraybackslash}p{5em}|} 
\hline
configuration & style layers & content layers  & $\lambda$ & initialization & input   alignement\\
\hline\hline
VI & $\ls^{\text{txtr}}$ & \path{conv4_2} & 20 & \multirow{4}{*}{\pbox{5.5em}{content \\ image}} & \multirow{4}{*}{yes}\\
\hhline{----~~}
VII & $\ls^{\text{txtr}} + \conv{5}$ & \path{conv4_2} & 20 & &\\
\hhline{----~~}
VIII &  $\ls^{\text{txtr}} + \conv{5}$ & \path{conv4_2} & 200 & &\\
\hhline{----~~}
V &  $\ls^{\text{txtr}} + \conv{5}$ & \path{conv5_2} & 200& &\\
\hline
\end{tabular}
\end{center}
\caption{Top left: part (e) of content image and a zoomed in patch of tail, top right: part (e) of style image and a zoomed in patch of tail; bottom row: zoomed in patch of tail in synthesized image at the same location as that in the top row corresponding to configurations VI, VII, VIII and V.}
\label{fig: Nr2_p3-2}
\end{figure}

Once the style-transfer algorithm is initialized with the content image, this strong prior on content limits the searching region of the algorithm in the image space. To ensure that a good style-transfered image is achievable, it's necessary to weaken the content constraint term in \eqref{eq: style-opt} and set $\lc$ to a deeper layer at a coarser scale than \path{conv4_2}. The second row of Fig.\ref{fig: Nr2_p3-2} compares results of different configurations with aligned input and content-based initialization. Configuration VI to VIII provide almost the same results (the first three columns) where the feather veins are more prominent than that generated by configuration V, our proposed content-aware style-transfer configuration (the last column).

\begin{figure}
\includegraphics[width = \textwidth]{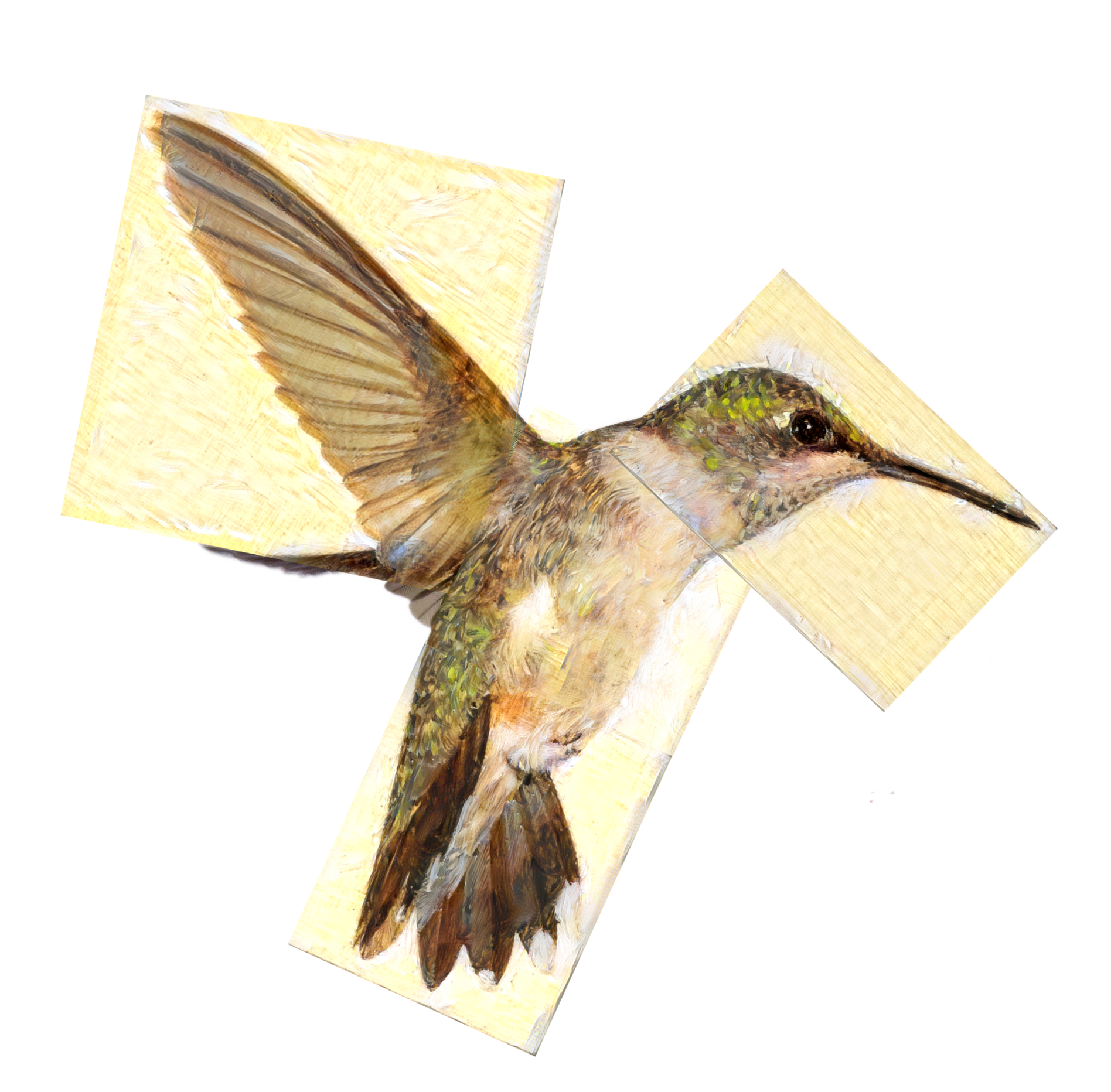}
\caption{stitched synthesized parts (a), (c), (d) and (e) of content image}
\label{fig: stitch-result}
\end{figure}
\begin{figure}
\includegraphics[width = \textwidth]{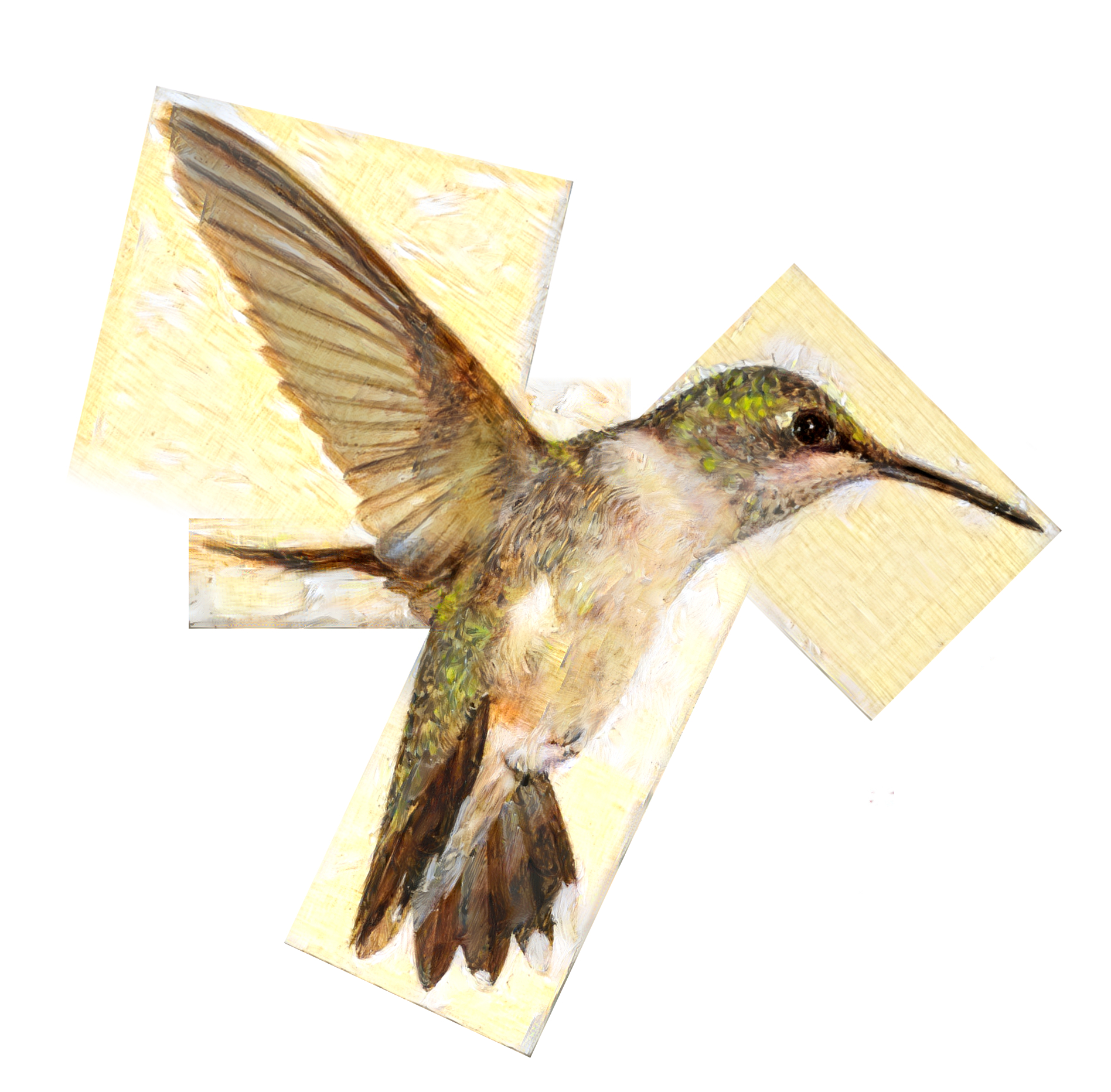}
\caption{merged synthesized parts (a), (b), (c), (d), (e) and (f) of content image}
\label{fig: merge-result}
\end{figure}

We apply the content-aware style transfer to pairs of content and style images for part (a) to (f). Fig.\ref{fig: stitch-result} shows an overlay of the synthesized result on part (a), (c), (d) and (e) up on the content photo. Fig.\ref{fig: merge-result} shows style transfer result on the whole object, which is obtained by overlaying the synthesized results of each part over the content photo and merging the overlapping regions between parts\footnote{The merging process is down in GIMP using the blending tool, which creates a mask on each part image with a smooth decay on the overlapping boundaries with other part images}.

There are some minor artifacts in the result shown in Fig.\ref{fig: merge-result}. The background color is not consistent within parts and there are white brush strokes in the background of part (a), (c), (d) and (f). Compare Fig.\ref{fig: merge-result} with the photo and the painting in Fig.\ref{fig: input}, the humming bird in the photo is slimmer than the painted bird who wears more white feathers, and the extra white feathers in the painting are redistributed to the background in the synthesized image. Although the white feathers is not in the photo, it is introduced by the style constraint to match the covariance of feature maps, resulting in a bias toward the object in the style image. The framework \eqref{eq: style-opt} does not achieve a strict separation of content and style information in a neural network. Futhermore, because of the spatial-independent property of the style constraint, there is no restriction where the white brush strokes could appear in the background. Next, we extend this algorithm for same-resolution content-aware style transfer to the super-resolution case. 

\subsection*{Super-resolution Style Transfer}

In the super-resolution case, we use the same photo and painting as previously discussed but downscale the photo to one-fourth of its original size. We apply the following iterative algorithm to generate a synthesized image of the same resolution of the style image:
\begin{enumerate}
\item[] {\it \hspace*{-2em} Input:} content image $I_c^0$ at scale $\alpha_0$, style image $I_s^{K}$ at scale $\alpha_K$
\item[{\it 1.}] downsize $I_s^K$ to $I_s^k$ at scale $\alpha_k, \, k < K,\, s.t. \,\alpha_{k-1}<\alpha_k<\alpha_{k+1}$  
\item[] {\it \hspace*{-1em} for $k = 0,\cdots,K$}
\item[] {\it 2.1} generate synthesized image $I_g^k$ transferring the style of $I_s^k$ to $I_c^k$ 
\item[] {\it 2.2} upscale $I_g^k$ to $\alpha_{k+1}$ and set as $I_c^{k+1}$ 
\item[] {\it \hspace*{-2.5em} Output:} synthesized image $I_g^K$ 
\end{enumerate}

\begin{sidewaysfigure}[ht]
\includegraphics[width = \textwidth]{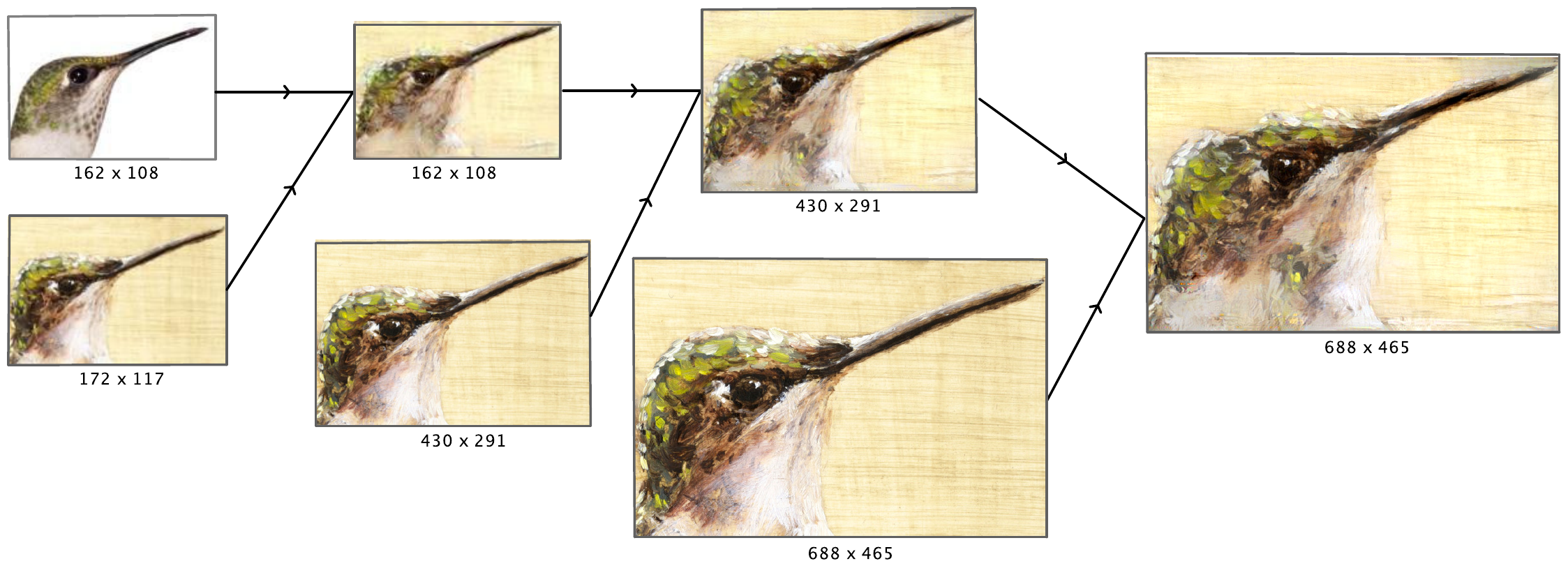}
\caption{super-resolution style transfer on reduced-size content photo with a series of painting increasing in size}
\label{fig: super-res}
\end{sidewaysfigure}

Fig.\ref{fig: super-res} shows the generation of $I_g^k$ along in the super-resolution algorithm. The final output is more similar to the painting than the synthesized image in the same-resolution case. In fact, the extra white feathers in the painting emerges gradually in $I_g^k$ as the scale increases.The iterative algorithm can regularize where to add the extra feature because the global structure of the object is captured and preserved through scales.
In the first iteration, the VGG-Net encodes global structure of the downsized input images $I_c^0,\,I_s^0$ at deep layers after several spatial pooling layers, and this structural information is kept in $I_g^0$ and passed on to the next iteration. In the $k^{th}$ iteration, the result of the last iteration is upscaled by a factor of $\alpha_k/\alpha_{k-1}$ and used as the new content image to generate $I_g^{k}$ in higher resolution. 
On the other hand, some fine content features such as gray spots on the neck of the bird diminishes along the iteration. The level of degradation in the $k^{th}$ iteration is proportional to $\alpha_k/\alpha_{k-1}$, or equivalently how underdetermined is the optimization.

\section*{Conclusion}
In this paper, we show that the formulation of the optimization \eqref{eq: style-opt} in \cite{Gatys2015c} doesn't account for spatial dependency of style patterns in the reference style image, and the basic style transfer algorithm is not content-aware, which is resolved by setting a strong initial prior on content while weaken the content constraint in the optimization algorithm. In addition, we show that it's necessary to align the input images to the same orientation, otherwise the induced content constraint and style constraint are not consistent as the VGG-Net is not rotation invariant. Furthermore, we extend this content-aware configuration into an iterative algorithm that transfers high resolution style to a low-resolution content image. Our proposed configuration for content-aware style transfer significantly improves the basic algorithm, and the extra features from the reference style image present in the synthesized image introduced by the style constraint means that the content and the style information is not completely separated by neural network.

\bibliography{ref}{}
\bibliographystyle{plain}
\end{document}